\documentclass[twocolumn]{galbot}

\usepackage{wrapfig}
\usepackage{tabularx}
\usepackage{textcomp}
\usepackage{stfloats}
\usepackage{balance}
\usepackage{url}
\usepackage{verbatim}
\usepackage{graphicx}
\usepackage{titlesec}
\usepackage{tocloft}
\usepackage{adjustbox}
\usepackage{multirow}
\usepackage{tikz}
\usepackage{comment}
\usepackage{amsmath,amssymb}
\usepackage{siunitx}
\usepackage{colortbl}
\usepackage{color}
\usepackage{booktabs} 
\usepackage{hyperref}
\usepackage{subcaption} 
\usepackage{arydshln}
\RequirePackage{xspace}
\makeatletter
\DeclareRobustCommand\onedot{\futurelet\@let@token\@onedot}
\def\@onedot{\ifx\@let@token.\else.\null\fi\xspace}

\makeatother

\usepackage{makecell}

\usepackage{pifont}
\usepackage{bbding}
\usepackage{fontawesome}
\usepackage{xspace}

\usepackage{float}

\newlength\savewidth

\newcolumntype{x}[1]{>{\centering\arraybackslash}p{#1pt}}
\newcolumntype{y}[1]{>{\raggedright\arraybackslash}p{#1pt}}
\newcolumntype{z}[1]{>{\raggedleft\arraybackslash}p{#1pt}}

\renewcommand{\paragraph}[1]{\vspace{1mm}\noindent\textbf{#1}}

\usepackage{xcolor}
\usepackage{array}
\usepackage{bbm}
\usepackage{collcell,xfp}
\usepackage{pgf}
\usepackage[most]{tcolorbox}
\usepackage{csquotes}
\usepackage[noorphans,vskip=1em,leftmargin=1em]{quoting}
\usepackage{enumitem} 
\usepackage{forest}
\usepackage{caption}
\usepackage{longtable}
\usepackage[T1]{fontenc}

\renewcommand{\paragraph}[1]{\vspace{1.25mm}\noindent\textbf{#1}}

\usepackage{algorithm}
\usepackage{listings}

\definecolor{codeblue}{rgb}{0.25, 0.5, 0.5}
\definecolor{codekw}{rgb}{0.35, 0.35, 0.75}
\lstdefinestyle{Pytorch}{
    language = Python,
    backgroundcolor = \color{white},
    basicstyle = \fontsize{9pt}{8pt}\selectfont\ttfamily\bfseries,
    columns = fullflexible,
    aboveskip=1pt,
    belowskip=1pt,
    breaklines = true,
    captionpos = b,
    commentstyle = \color{codeblue},
    keywordstyle = \color{codekw},
}

\definecolor{green}{HTML}{009000}
\definecolor{red}{HTML}{ea4335}

\definecolor{linecolor1}{RGB}{246, 248, 239}
\definecolor{linecolor2}{RGB}{230, 234, 217}
\definecolor{linecolor3}{RGB}{211, 222, 190}

\DeclareRobustCommand{\ours}{HumanTracker\xspace}

\title{\ours: Towards Comprehensive and Human-Aligned Motion Tracking Benchmark}

\author[1, 3, *]{Dairu Liu}
\author[2, 3, *]{Zekun Qi}
\author[3, *]{Jiayu Zeng}
\author[2, 3]{Ruixi Yu}
\author[2, 3]{Yu Guan}
\author[3]{Yintianrun Zhang}
\author[2, 3]{Xuchuan Chen}
\author[3, 4]{Sikai Liang}
\author[2, 3]{Zekai Li}
\author[3]{Chenghuai Lin}
\author[3]{Xinqiang Yu}
\author[4]{Wenyao Zhang}
\author[3, 5, \dagger]{He Wang}
\author[2, 3, 6, \dagger]{Li Yi}

\affiliation[1]{Nankai University}
\affiliation[2]{Tsinghua University}
\affiliation[3]{Galbot}
\affiliation[4]{Shanghai Jiao Tong University}
\affiliation[5]{Peking University}
\affiliation[6]{Shanghai Qi Zhi Institute}

\contribution[*]{Equal Contribution}
\contribution[\dagger]{Corresponding author}

\date{\today}
\page{\url{https://dairuliu.github.io/humantracker}}
\code{\url{https://github.com/GalaxyGeneralRobotics/HumanTracker}}

\abstract{
Humanoid motion tracking is central to teleoperation and whole-body imitation, yet evaluation often disagrees with what people perceive in videos.
Kinematic errors average per-frame pose differences but miss the physical artifacts that matter most, particularly unstable support and incorrect contacts such as foot skating and mistimed touch-downs.
Meanwhile, widely used test suites are small and lack the diversity needed to stress contact-rich, long-horizon behaviors.
We introduce \ours to make humanoid tracking evaluation both perceptually aligned and scalable.
The \ours{} benchmark contains approximately 153 hours of optical motion trajectories from multiple professional performers, organized into four motion families with text labels for fine-grained diagnosis.
We further propose HumanScore, a preference-aligned metric trained on 12K motion pairs containing 24K motions.
Across representative state-of-the-art trackers, HumanScore better predicts human preferences and reveals contact and stability failures that kinematic metrics often miss.
}

\makeatletter
\apptocmd{\mymaketitle}{%
  \begin{center}
      \captionsetup{type=figure}
      \vspace{12pt}
      \includegraphics[width=\textwidth]{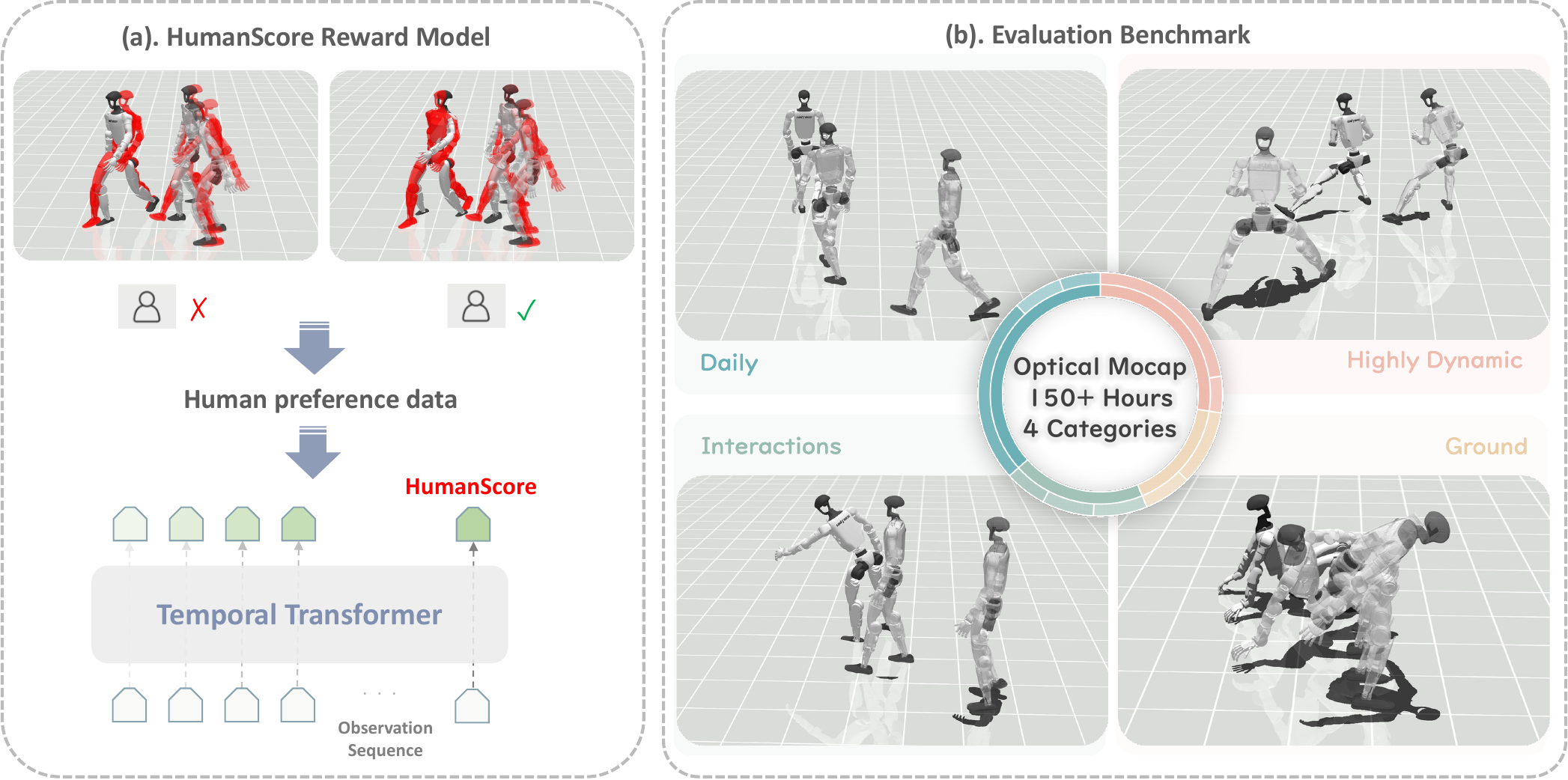}
      \vspace{2pt}
      \captionof{figure}{\textbf{\ours overview}. Left: we train a Human-Aligned Reward Model on pairwise preference data from tracking rollouts, producing a scalar HumanScore via a temporal Transformer. Right: the \ours Benchmark provides 153 hours of motion trajectories across four families: daily tasks, highly dynamic motions, interaction motions, and ground-level movements.}
      \label{fig:teaser}
  \end{center}
}{}{}
\makeatother

\begin{document}
\maketitle
\pagestyle{empty}

\section{Introduction}
\label{sec:intro}

Motion tracking is becoming the cornerstone of humanoid robot control~\cite{phc23,gmt25,sonic25,unitracker25,beyondmimic25,humanoid_ntp25,kungfubot25}. This ability drives applications in teleoperation and whole-body imitation~\cite{twist25,twist2_25,omnih2o24,humanplus24,clone25,h2o24,mhc25}. Most systems track a reference motion using a learned feedback policy in physics simulation~\cite{phc23,gmt25,sonic25,unitracker25}. Yet, measuring the true quality of this tracking remains surprisingly difficult. Two major issues currently limit progress in this field.

The first major issue is how we measure success. Motion tracking looks easy to measure: one can compare the rollout pose to the reference pose and report kinematic errors such as mean joint error and key point error~\cite{gmt25,sonic25,unitracker25,any2track25,humanoidgpt26}. However, we find a clear gap between these numbers and what people see in videos. A rollout may achieve low kinematic error but still look bad, especially under contacts~\cite{phuma25,omniretarget25,gmr25,ictrl25}. Feet may slide on the ground, and contacts may break and reattach at the wrong time; these artifacts are exactly what contact-aware global motion reconstruction and refinement methods target in video-based settings as well~\cite{wham24,proxycap24,rohm24}. These artifacts are the key aspect to decide whether the motion looks stable, smooth and human-like.

This gap is not a corner case but a consequence of what kinematic metrics measure. By treating tracking as a per-frame pose matching problem and averaging errors over joints and time, these metrics fail to capture contact, support, balance, and the accumulation of errors in closed-loop control. As a result, two rollouts can have similar pose errors yet differ substantially in stability. Observers reliably prefer the rollout with clean contacts and steady support, even when MPJPE is similar. Figure~\ref{fig:teaser} highlights this mismatch.

The second major issue is the scope of evaluation data. Despite the availability of large motion repositories such as PHUMA~\cite{phuma25} and SONIC~\cite{sonic25}, the most commonly used evaluation suite for humanoid tracking is still an AMASS test set with only 140 sequences~\cite{amass19}. This small set lacks diversity and under-represents the long tail of human movement, including challenging contact transitions, asymmetric balancing, and complex recoveries. Moreover, results are often summarized as a single aggregate score, without a detailed breakdown by motion category, making it difficult to pinpoint exactly where and why a tracker fails. Table~\ref{tab:dataset_comparison} compares \ours{} with representative large-scale motion datasets in scale, categorization and text annotation.

\begin{table}[h]
\centering
\caption{\textbf{Comparison of large-scale humanoid motion datasets.} \ours uniquely provides approximately 153 hours of motion trajectories explicitly organized into distinct motion categories for comprehensive evaluation.}
\label{tab:dataset_comparison}
\small
\resizebox{1.0\linewidth}{!}{
\setlength{\tabcolsep}{3.5pt}
\begin{tabular}{lcccc}
\toprule
\textbf{Dataset} & \textbf{Clips} & \textbf{Hours} & \textbf{Categories} & \textbf{Text Label} \\
\midrule
AMASS~\cite{amass19} & $>$11K & $>$40 & No & No \\
HumanML3D~\cite{humanml3d22} & 14.6K & 28.6 & No & Yes \\
PHUMA~\cite{phuma25} & 76K & 73 & No & No \\
\textbf{\ours} & 25K & \textbf{153} & \textbf{4} & \textbf{Yes} \\
\bottomrule
\end{tabular}
}
\end{table}

We introduce \ours to address these problems. To address the metric misalignment, we develop a preference-aligned evaluation: we collect human comparisons on synchronized tracking videos and train a reward model to predict these preferences~\cite{RLHF17,InstructGPT22,IterativeRLHF24}. We call this metric HumanScore. Unlike joint error, HumanScore explicitly targets human preferences, perceptual stability, and realistic physical contacts. It penalizes the exact physical artifacts that look wrong to human observers. Importantly, we demonstrate that HumanScore captures nuanced perceptual qualities that cannot be simply reduced to rule-based diagnostics such as foot slip or root drift. Extensive experiments and visual analyses demonstrate the accuracy and zero-shot generalization of HumanScore.

To address the lack of diversity in existing test suites, we provide approximately 153 hours of optical motion trajectories paired with category and text labels, complementing existing large-scale motion sources~\cite{amass19,motionx23,motionx++25,phuma25}. All source motions are recorded in a controlled studio with a multi-camera optical system by 24 professional performers, including dance teachers and fitness coaches, yielding high-fidelity references for contact-rich tracking evaluation. We explicitly organize the dataset into four distinct families covering daily tasks, highly dynamic movements, interaction motions, and ground-level motions, enabling per-family metrics and fine-grained diagnosis of failure modes. \ours substantially exceeds prior mocap datasets used for humanoid tracking. We comprehensively evaluate several state-of-the-art trackers, including GMT, TWIST2, SONIC, and Humanoid-GPT, under our benchmark~\cite{gmt25,twist2_25,sonic25,humanoidgpt26}.

In summary, our main contributions are threefold. First, we highlight the failure of traditional kinematic metrics and introduce HumanScore to align evaluation with humans. Second, we provide a massive motion tracking benchmark featuring approximately 153 hours of diverse, categorized optical trajectories. Third, we establish a rigorous and standardized evaluation protocol to ensure future progress in humanoid tracking is both measurable and meaningful.

\begin{figure*}[t]
\centering
\includegraphics[width=0.99\linewidth]{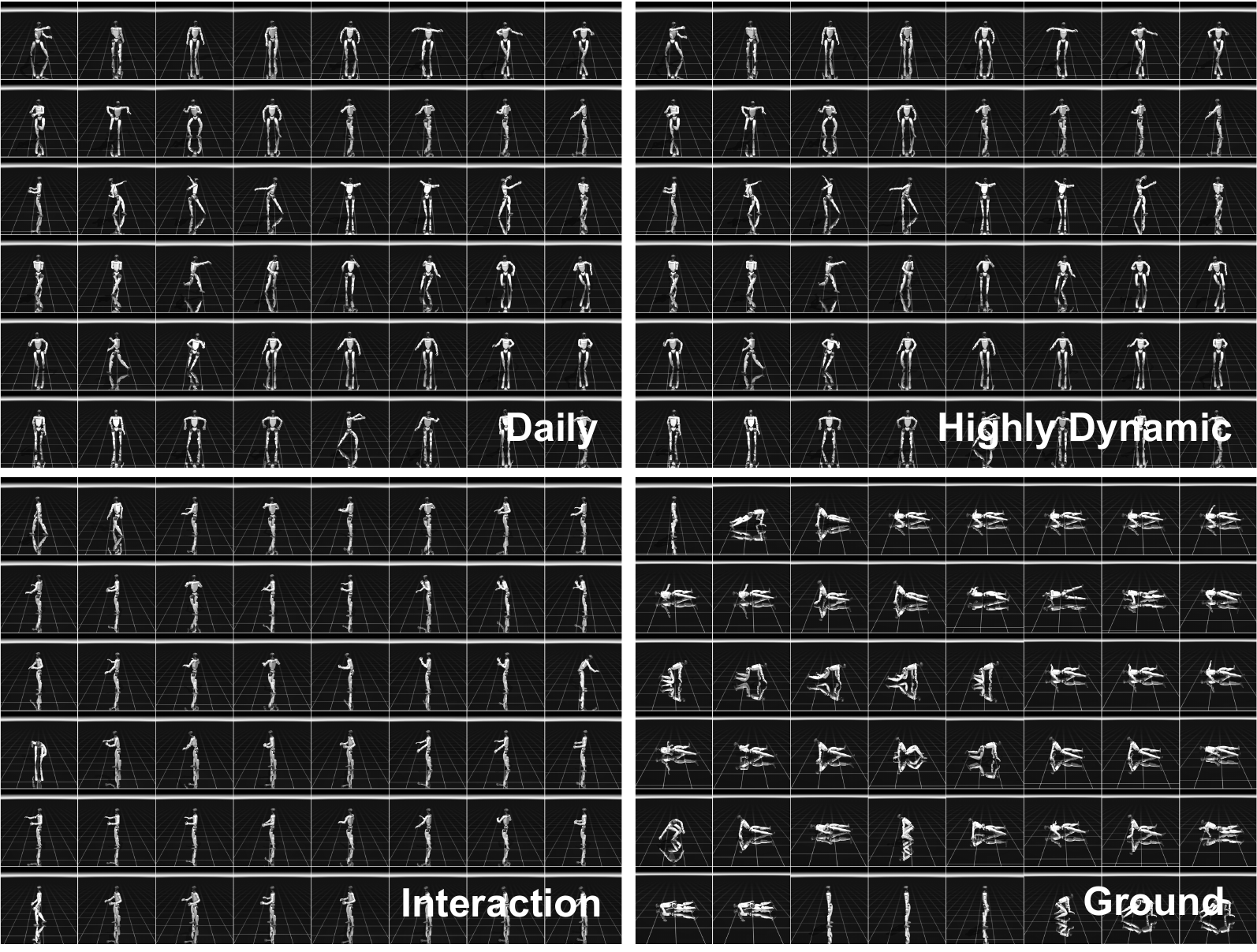}
\caption{\textbf{Motion taxonomy.} \ours{} groups motions into four families according to their dominant tracking and contact regimes. The taxonomy supports category-level diagnosis rather than only a single aggregate score.}
\label{fig:taxonomy}
\end{figure*}

\section{Related Work}
\label{sec:related_works}

\subsection{Humanoid motion tracking}
DeepMimic~\cite{deepmimic18} established reference conditioned policy learning for physics-based character control. PHC and UHM~\cite{phc23,uhm23} extended this paradigm to diverse humanoid motion. Recent systems broaden motion coverage and control robustness through several mechanisms. GMT, SONIC, Humanoid-GPT and UniTracker~\cite{gmt25,sonic25,humanoidgpt26,unitracker25} emphasize scale and general tracking. ResMimic, MHC and iCTRL~\cite{resmimic25,mhc25,ictrl25} explore residual correction, multimodal commands and constrained control. Any2Track~\cite{any2track25} adapts tracking policies to changes in terrain, external forces and physical properties. HumanoidPF~\cite{xue2026collisionfreehumanoidtraversalcluttered} uses a humanoid potential field to learn collision avoidance skills for cluttered indoor scenes. Adversarial Differential Discriminators~\cite{add25} provide a learned motion imitation objective. OmniTrack~\cite{omnitrack26} constructs physically consistent references, while LIMMT~\cite{guan2026limmtmotiontracking} studies how carefully selected training motions can improve tracking. The same capability underpins whole-body teleoperation and imitation systems, including TWIST, OmniH2O, HumanPlus, CLONE and H2O~\cite{twist25,omnih2o24,humanplus24,clone25,h2o24}. These advances have produced increasingly capable trackers, but their reported performance is difficult to compare because results depend not only on the learned policy, but also on the reference set, simulator, action representation, initialization, and termination rule. \ours{} therefore treats evaluation as a controlled experiment. Tracker specific policy interfaces are preserved, whereas the reference representation, rollout accounting, and reported metrics are standardized.

\subsection{Motion data evaluation}
Large motion repositories such as AMASS, Motion-X and Motion-X++~\cite{amass19,motionx23,motionx++25} have substantially expanded the diversity of human motion available for learning. Recent works~\cite{humanml3d22,freemotion24,omomo23,latent26} further support semantic retrieval and conditional generation. For humanoid tracking, however, dataset size alone does not define a useful benchmark. Human motion must also be retargeted to the robot morphology, remain physically plausible around contacts, and contain sufficiently difficult regimes to expose controller failures. PHUMA, OmniRetarget and GMR~\cite{phuma25,omniretarget25,gmr25} address these requirements through physically grounded data or robot retargeting. WHAM, ProxyCap and RoHM~\cite{wham24,proxycap24,rohm24} address related errors in global human motion reconstruction. Switch-JustDance~\cite{switch25} provides a complementary benchmark based on whole-body skills from a commercial motion game. Existing tracking evaluations nevertheless remain concentrated on comparatively small test sets and often collapse heterogeneous motions into one aggregate value. \ours{} complements prior motion sources with a large, explicitly categorized test bed whose four families separate steady locomotion, rapid impact rich motion, interaction motions, and ground level transitions with multiple contacts.

\begin{table*}[t]
\centering
\caption{\textbf{\ours{} dataset statistics.} The benchmark contains approximately 153 hours and 25K clips across four complementary tracking regimes.}
\label{tab:stats}
\setlength{\tabcolsep}{8.5pt}
\resizebox{0.77\linewidth}{!}{
\begin{tabular}{lccc}
\toprule
\textbf{Family} & \textbf{Hours} & \textbf{\#Clips} & \textbf{Typical challenges} \\
\midrule
Daily & 89 & 9.7k & steady locomotion, mild contacts \\
Highly Dynamic & 11 & 2.7k & impacts, aerial phases, fast footwork \\
Interaction & 48 & 10.9k & human-like, stable, smooth hands-body coordination\\
Ground & 5 & 1.6k & low posture, multi-contact transitions \\
\midrule
Total & 153 & 25K & diverse \\
\bottomrule
\end{tabular}
}
\end{table*}

\section{\ours: Benchmark and Preference-Aligned Evaluation}
\label{sec:benchmark}


\subsection{The \ours{} benchmark}
\label{sec:dataset}

\paragraph{Motion collection and processing.}
The released \ours{} benchmark contains approximately 153 hours of optical motion trajectories from 24 professional performers. The performers include dance teachers, fitness coaches, tennis coaches and full-time motion-capture actors. The performers and recording plan were chosen to cover both routine movement and motions that place substantially different demands on a humanoid controller. We retarget each fitted human motion to the benchmark humanoid using General Motion Retargeting (GMR)~\cite{gmr25}. Because a visually valid human recording is not automatically a valid robot reference, we inspect the retargeted sequences and remove segments with capture or processing artifacts such as unexplained floating, ground penetration and discontinuous contacts. Each released clip contains a top-level motion-family label, a natural-language description, a fitted SMPL sequence~\cite{smpl15} and a robot-space reference trajectory in \texttt{qpos} format. These representations support semantic subset selection while keeping the input to every evaluated tracker identical. Figure~\ref{fig:taxonomy} illustrates the resulting motion coverage, and Table~\ref{tab:stats} summarizes the family-level scale of the benchmark.

\paragraph{A taxonomy for diagnostic evaluation.}
The four motion families are defined by the failure regimes that they expose. \textit{Daily} contains walking, turning and routine gestures, and therefore measures steady-state stability and residual drift under comparatively regular support. \textit{Highly Dynamic} contains jumps, kicks, acrobatics and fast dance footwork, for which impacts and rapid support switching amplify phase and timing errors. \textit{Interaction} contains human body motions associated with actions involving objects or the surrounding environment. It evaluates hand, arm and whole body coordination in the human reference. These trajectories can also provide kinematic priors for humanoid manipulation. \textit{Ground} covers kneeling, sitting, rolling and recovery, where a low centre of mass and multiple simultaneous contacts make the controller sensitive to contact geometry and friction. The distribution intentionally reflects the frequency of the captured activities rather than forcing equal family sizes; all benchmark results are therefore reported by family as well as in aggregate. The complete dataset is split 9:1 into disjoint training and test partitions, with the family distribution preserved and duplicate motions kept within one partition.

\begin{figure*}[t]
\centering
\includegraphics[width=0.98\linewidth]{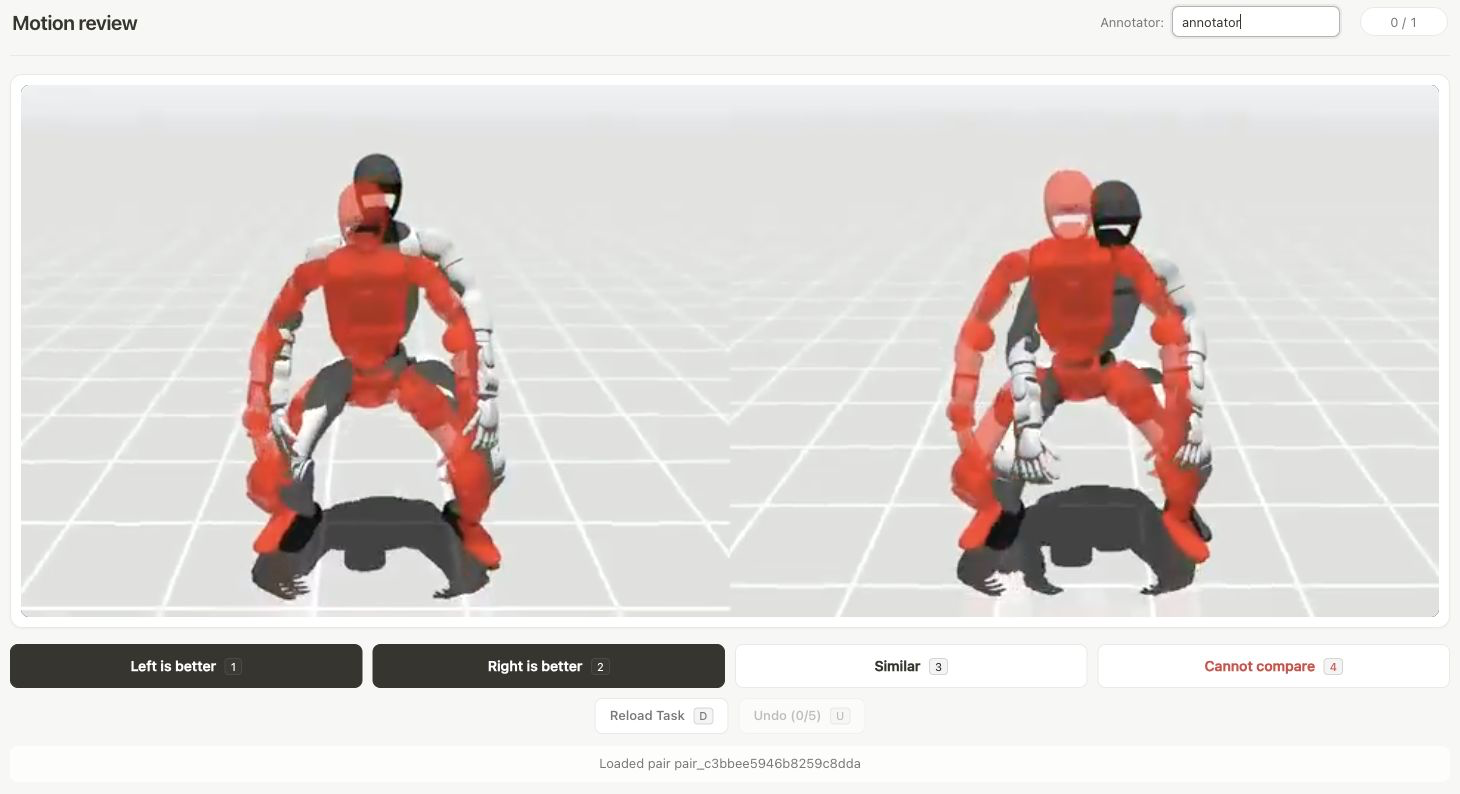}
\caption{\textbf{Preference collection interface.} Annotators view two synchronized rollouts of the same reference segment. Display order is randomized and the available responses are \emph{Left is better}, \emph{Right is better}, \emph{Similar} and \emph{Cannot compare}.}
\label{fig:pref_ui}
\end{figure*}

\subsection{Standardized tracker evaluation}
\label{sec:protocol}

For every method, we convert the reference to the same 29-DoF humanoid \texttt{qpos} representation and execute the policy through a common MuJoCo evaluation entry point. Each tracker retains its native policy observations and action decoder, and the evaluator instead standardizes the motion list, robot model, reference indexing, rollout accounting and metric implementation. The control trajectory is recorded at 50\,Hz. At every step, the evaluator stores the simulated generalized position and velocity, policy action and motor target, foot contacts and contact forces, foot and pelvis velocities, and 14 keypoint poses and spatial velocities. The same state history is used for conventional diagnostics and HumanScore, which prevents differences in post-processing from being mistaken for differences between trackers.

We use the same tracking metric and success criterion with SONIC~\cite{sonic25} for each tracker. It measures vertical position error at the pelvis, both ankles and both wrists, together with pelvis rotation error. The episode fails when any vertical error exceeds 0.25\,m, the pelvis rotation error exceeds 1\,rad, or the generalized position or velocity contains a nonfinite value. We report \emph{Succ} as the fraction of completed episodes and \emph{MPJPE} as the mean absolute error over the 29 actuated joint angles, in radians, over the executed portion of each rollout. Additional diagnostics include joint-velocity error, keypoint-position error, foot-contact agreement and finite-difference joint acceleration and jerk. All reported benchmark values use the \ours{} test split.

\subsection{Preference data construction}
\label{sec:pref}

\paragraph{Rollout segmentation for paired comparison.}
The preference pool is generated exclusively from motions in the \ours{} training split, so no benchmark test motion is used to train HumanScore. For each source motion, GMT~\cite{gmt25}, Humanoid-GPT~\cite{humanoidgpt26}, SONIC~\cite{sonic25} and TWIST2~\cite{twist2_25} produce aligned rollouts of the same robot-space reference. At 50\,Hz, every rollout is divided into consecutive 250-frame windows, each spanning 5\,s. If a rollout ends with a shorter window, the final short window is retained.

\paragraph{Uniform rollout pair sampling.}
We concatenate all aligned windows in a deterministic order by motion family, source motion and temporal position, and assign each window a global index. From this catalogue we select a uniformly spaced set of unique indices over the full ordered list. This construction samples the complete \ours{} training distribution without manually favouring visually difficult examples or any particular family. Each selected window yields exactly one comparison between two of the four trackers. The six unordered tracker combinations are allocated equally, so every pairing contributes the same number of comparisons and each tracker appears equally often. Candidate order is alternated within each combination and the final task order is deterministically shuffled, so no tracker is favoured by display position. The resulting preference pool is partitioned into train and test sets.

\paragraph{Human annotation and split.}
The annotation panel comprised six doctoral researchers specializing in humanoid robotics, providing domain expert judgments of balance, contact, tracking stability and motion naturalness. Using the interface shown in Figure~\ref{fig:pref_ui}, they first decide whether either rollout fails to complete the motion or loses balance; when both remain viable, they compare jitter, foot sliding, locomotion consistency and whole-body naturalness in that order. They choose \emph{Similar} when neither candidate is meaningfully better and \emph{Cannot compare} when both are unusable or the evidence is insufficient. The interface randomizes display order and records the underlying candidate indices, eliminating a fixed association between display side and tracker identity. The collected labels comprise strict preferences, similar pairs and cannot-compare pairs. Cannot-compare pairs are excluded from reward-model optimization. We construct an 80/20 training/test split by grouping records by the original \texttt{motion\_id}; all clips from one source motion therefore remain in one partition. The split is balanced jointly over family, tracker pairing, label type, annotator and the remaining records define the training and test preference sets.

\begin{figure*}[t]
\centering
\includegraphics[width=1.0\linewidth]{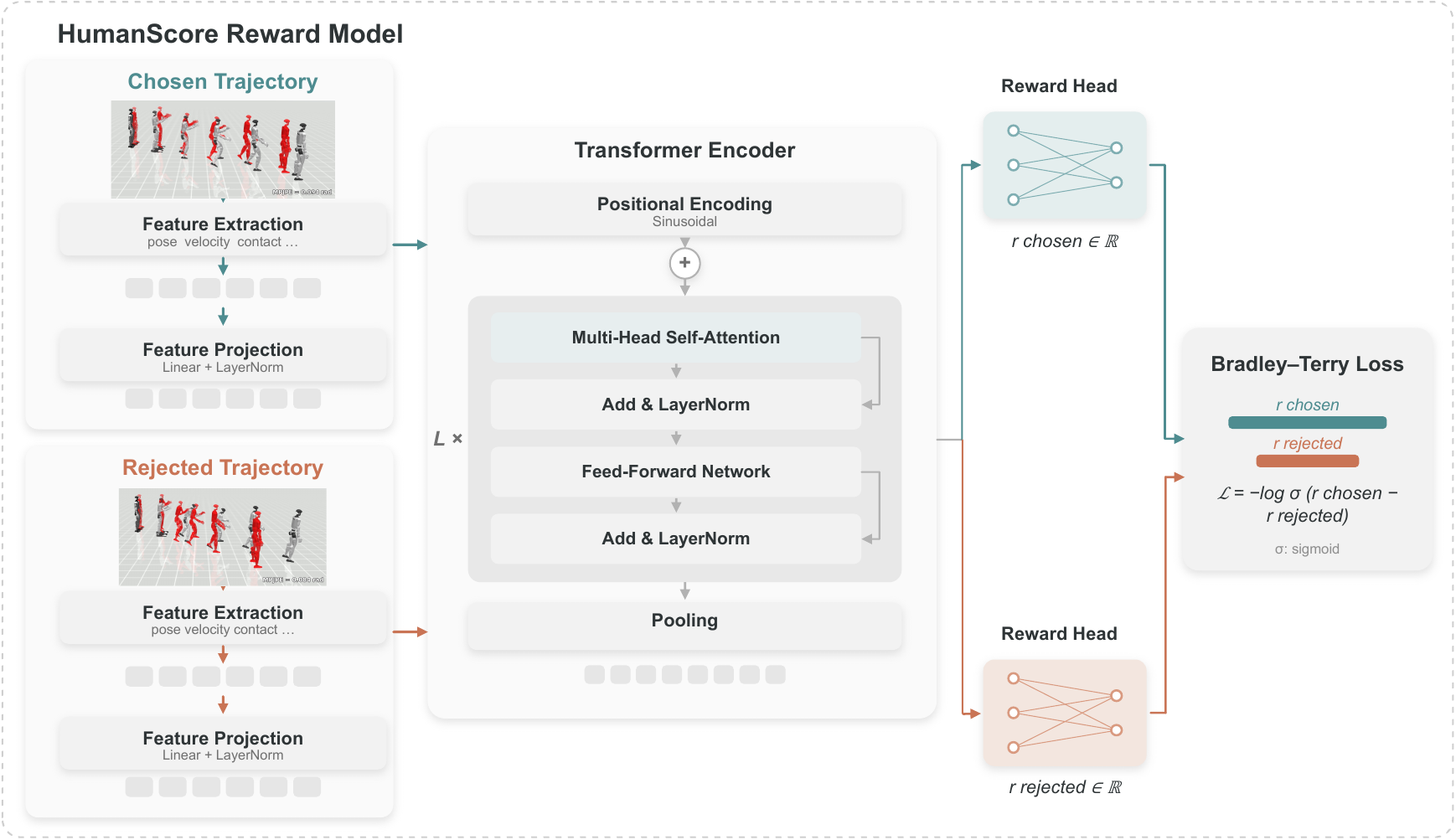}
\caption{\textbf{HumanScore trajectory reward model.} Current reference and simulated state features form one token per frame. A temporal Transformer processes the valid tokens, and masked mean pooling forms a trajectory representation. The diagram shows the Bradley–Terry loss on the unbounded rewards $r_{\mathrm{chosen}}$ and $r_{\mathrm{rejected}}$ for a strict comparison.}
\label{fig:reward_model}
\end{figure*}

\subsection{HumanScore}
\label{sec:reward}

\paragraph{Trajectory representation.}
Figure~\ref{fig:reward_model} summarizes how the reward model maps a rollout segment $\tau$ to an unbounded reward $r_\theta(\tau)\in\mathbb{R}$. Although human preferences are collected from rendered videos, the model operates directly on simulator trajectories, avoiding dependence on camera viewpoint or rendering choices. Each frame is represented by a 539-dimensional vector. It contains 70 dimensions for the current reference state and 469 dimensions for simulated state, control, measured contact dynamics, root motion and current keypoint kinematics. The reported model does not use future reference residuals. Conditioning on the current reference and rollout allows HumanScore to assess tracking quality rather than motion plausibility alone. The complete feature decomposition is provided in the appendix.

The frame vector is linearly projected, normalized, and augmented with sinusoidal positional encoding before being processed by a Transformer encoder~\cite{Transformer17}. A padding mask is applied in every attention layer. The elementwise mean of the valid output tokens then forms the trajectory representation, which an MLP maps to the scalar reward. The same mask is used for the retained tail clips. Segments shorter than 250 frames are padded with zeros on the right, while the validity mask excludes padded positions from both attention and temporal pooling. Thus, full and truncated windows can be processed by the same model without introducing padding artifacts.

\paragraph{Preference objective.}
For a strict pair $i\in\mathcal D$, the model assigns rewards $r_{\mathrm{chosen}}^{(i)}$ and $r_{\mathrm{rejected}}^{(i)}$ to the chosen and rejected trajectories. We define $\Delta_i=r_{\mathrm{chosen}}^{(i)}-r_{\mathrm{rejected}}^{(i)}$ and use the Bradley–Terry loss~\cite{bradleyTerry52}
\[
\mathcal L_{\mathrm{diff}}^{(i)}=-\log\sigma(\Delta_i).
\]
The less frequent \emph{Similar} pairs provide an equality constraint. For a Similar pair $j\in\mathcal S$, let $r_a^{(j)}$ and $r_b^{(j)}$ denote the rewards of its two trajectories, and define $\Delta_j=r_a^{(j)}-r_b^{(j)}$. Their symmetric loss is
\[
\mathcal L_{\mathrm{similar}}^{(j)}
=-\frac{1}{2}\log\sigma(\Delta_j)
-\frac{1}{2}\log\sigma(-\Delta_j).
\]
Every pair contributes once to the batch objective
\[
\mathcal L=\frac{1}{|\mathcal D|+|\mathcal S|}
\left(
\sum_{i\in\mathcal D}\mathcal L_{\mathrm{diff}}^{(i)}
+\sum_{j\in\mathcal S}\mathcal L_{\mathrm{similar}}^{(j)}
\right).
\]

\begin{table*}[t]
\centering
\caption{\textbf{Zero-shot evaluation on \ours{}.} All trackers are evaluated without training or fine-tuning on the test set. We report completion rate under the whole-body termination criterion (Succ), mean absolute joint-angle error over 29 actuated joints (MPJPE), and perceptual trajectory quality (HumanScore, 0 to 100). Higher Succ and HumanScore and lower MPJPE indicate better performance; bold denotes the best result within each motion family.}
\label{tab:main_zeroshot}
\scriptsize
\setlength{\tabcolsep}{1.5pt}
\renewcommand{\arraystretch}{1.05}
\resizebox{1.0\linewidth}{!}{
\begin{tabular}{lccc ccc ccc ccc}
\toprule
& \multicolumn{3}{c}{\textbf{Daily}} & \multicolumn{3}{c}{\textbf{Highly Dynamic}} & \multicolumn{3}{c}{\textbf{Interaction}} & \multicolumn{3}{c}{\textbf{Ground}} \\
\cmidrule(lr){2-4}\cmidrule(lr){5-7}\cmidrule(lr){8-10}\cmidrule(lr){11-13}
\multicolumn{1}{c}{\textbf{Method}} & \multicolumn{12}{c}{} \\[-1.55ex]
& \shortstack{Succ\\(\%)} & \shortstack{MPJPE\\(rad)} & \shortstack{Human\\Score}
& \shortstack{Succ\\(\%)} & \shortstack{MPJPE\\(rad)} & \shortstack{Human\\Score}
& \shortstack{Succ\\(\%)} & \shortstack{MPJPE\\(rad)} & \shortstack{Human\\Score}
& \shortstack{Succ\\(\%)} & \shortstack{MPJPE\\(rad)} & \shortstack{Human\\Score} \\
\midrule
GMT~\cite{gmt25} & 17.0 & 0.250 & 2.4 & 36.2 & 0.196 & 7.0 & 81.4 & 0.205 & 11.7 & 0.0 & 0.456 & 4.0 \\
TWIST2~\cite{twist2_25} & 60.1 & 0.105 & 10.1 & 39.9 & 0.112 & 16.9 & 91.3 & 0.111 & 28.3 & 0.0 & 0.341 & 4.5 \\
SONIC~\cite{sonic25} & 93.8 & 0.102 & 49.5 & 82.1 & 0.118 & 41.0 & \textbf{97.6} & 0.128 & 54.6 & 20.1 & 0.231 & \textbf{26.5} \\
Humanoid-GPT~\cite{humanoidgpt26} & \textbf{94.4} & \textbf{0.046} & \textbf{54.7} & \textbf{86.9} & \textbf{0.047} & \textbf{49.2} & 97.2 & \textbf{0.070} & \textbf{56.8} & \textbf{32.9} & \textbf{0.216} & 24.9 \\
\bottomrule
\end{tabular}
}
\end{table*}

\subsection{HumanScore computation}
\label{sec:release}

At evaluation time, a rollout $\tau$ of $F$ frames is divided into $N=\lceil F/250\rceil$ consecutive windows. Let $L_i\leq250$ be the number of actual frames in window $s_i$, so $\sum_i L_i=F$. A short final window is padded on the right and evaluated with its validity mask. For reporting, the unbounded reward of each window is mapped to a bounded reward
\[
\rho_\theta(s_i)=\sigma\!\left(r_\theta(s_i)\right)\in(0,1).
\]
We then average the window rewards according to the number of frames that they represent
\[
\mathrm{HumanScore}(\tau)
=\frac{100}{F}\sum_{i=1}^{N}L_i\rho_\theta(s_i).
\]
Padding is used only to form the model input and contributes no weight to the average.
Under the sigmoid mapping used at inference, \(\rho_\theta(s_i)\) ranges from \(4\times10^{-8}\) to \(0.99\) over the training data distribution, providing sufficient dynamic range to distinguish trajectories according to human preferences.

\section{Experiments}
\label{sec:experiments}


\subsection{Experimental setup}

\paragraph{Evaluation setting.}
We evaluate GMT, TWIST2, SONIC and Humanoid-GPT through the standardized protocol of Sec.~\ref{sec:protocol}. Each policy retains its native observation and action-processing stack, while all methods receive the same retargeted reference motions and are measured by the same evaluator. We apply the whole-body metric defined in Sec.~\ref{sec:protocol} to every tracker. None of the trackers is trained or fine-tuned on \ours{}. Every benchmark number is computed on the \ours{} test split and is reported separately for \textit{Daily}, \textit{Highly Dynamic}, \textit{Interaction} and \textit{Ground}.

\paragraph{Metrics.}
We report three complementary measures of tracking quality. Succ captures catastrophic loss of tracking but cannot distinguish the quality of two completed rollouts. MPJPE measures local reference fidelity but averages over time and joints. HumanScore measures trajectory-level preference and is designed to respond to contact, stability and smoothness as well as pose. We evaluate HumanScore in consecutive 5-second windows and retain a final shorter window using right-zero padding and a validity mask. For the preference study, we additionally compare joint-velocity error, keypoint-position error, foot-contact agreement, and mean joint acceleration.

\begin{figure*}[t]
\centering
\begin{subfigure}[t]{0.495\linewidth}
\centering
\includegraphics[width=\linewidth]{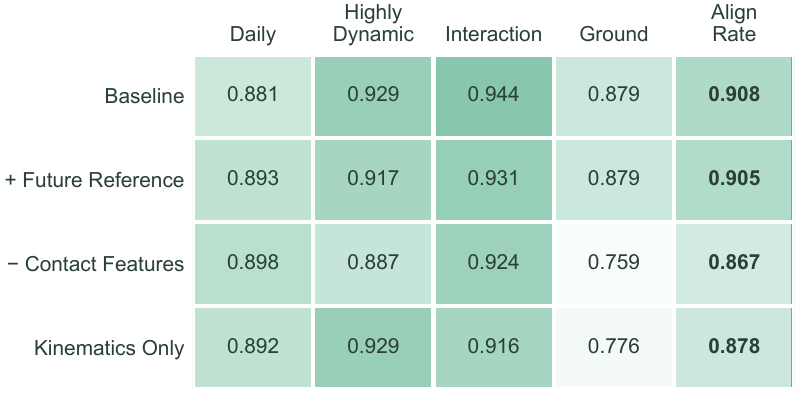}
\caption{Models trained with different input features.}
\end{subfigure}
\hfill
\begin{subfigure}[t]{0.495\linewidth}
\centering
\includegraphics[width=\linewidth]{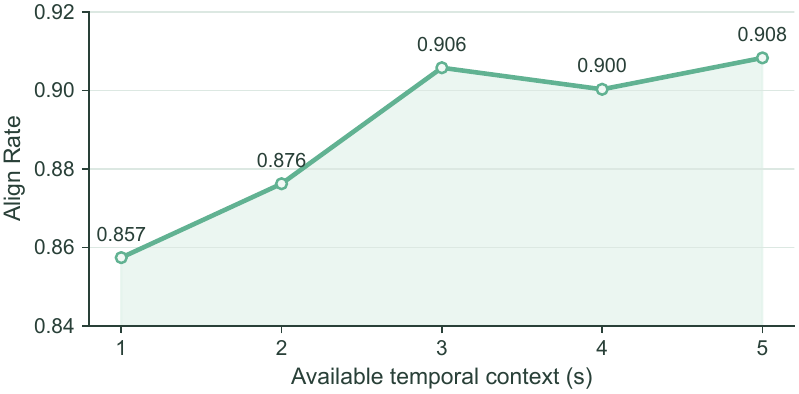}
\caption{Baseline with restricted temporal context.}
\end{subfigure}
\caption{\textbf{HumanScore sensitivity.} Align Rate is computed within each motion family and then averaged across 4 families. Panel a evaluates different input feature sets. Panel b restricts the temporal prefix available to the baseline at inference.}
\label{fig:rm_analysis}
\end{figure*}

\begin{table}[t]
\centering
\small
\caption{\textbf{Alignment with human preferences.} Align Rate is first computed within each motion family on strict test samples and then averaged across \textit{Daily}, \textit{Highly Dynamic}, \textit{Interaction} and \textit{Ground}. Each family therefore contributes one quarter of the reported value.}
\setlength{\tabcolsep}{7.5pt}
\begin{tabular}{lc}
\toprule
\textbf{Metric} & \textbf{Align Rate} \\
\midrule
HumanScore & \textbf{0.9083} \\
MPJPE (rad) & 0.8049 \\
MPJVE (rad/s) & 0.8404 \\
KPT Position MAE (m) & 0.8405 \\
Foot Contact Accuracy & 0.7882 \\
Avg Joint Accel (rad/s$^2$) & 0.6933 \\
Avg Joint Jerk (rad/s$^3$) & 0.7232 \\
\bottomrule
\end{tabular}
\label{tab:preference_alignment}
\end{table}

\subsection{Results}
\label{sec:benchmark_results}

Table~\ref{tab:main_zeroshot} shows that Humanoid-GPT is the strongest overall tracker, leading most comparisons and all three metrics on \textit{Daily} and \textit{Highly Dynamic}. SONIC is the closest competitor. It achieves the highest completion rate on \textit{Interaction} and the highest HumanScore on \textit{Ground}. These exceptions reveal a meaningful difference between the two trackers. Humanoid-GPT generally completes motions more reliably and follows the reference more closely, whereas SONIC produces \textit{Ground} rollouts that are perceived as more natural and stable. The family-level results therefore distinguish consistent overall tracking from strengths that emerge under a particular evaluation criterion. Moreover, it also shows that traditional metrics are not always aligned with HumanScore.

\subsection{Human preference comparison}
\label{sec:humanscore_eval}

Table~\ref{tab:preference_alignment} shows that HumanScore agrees with human preferences more consistently than any individual analytic diagnostic. Conventional metrics isolate pose, velocity or contact fidelity, whereas annotators also consider human-like, smoothness, stability and how errors develop over time. Their lower agreement indicates that no single diagnostic captures the full trajectory quality reflected in human judgments.

The comparison also rules out several distributional shortcuts. Grouping the test set by source motion prevents adjacent clips from the same sequence from crossing partitions. Averaging agreement within each family prevents the more frequent \textit{Daily} and \textit{Interaction} samples from obscuring performance on \textit{Ground} and \textit{Highly Dynamic}. Tracker pairings are balanced by construction, so the result cannot be explained by a dominant family or tracker matchup.

\subsection{Sensitivity Analysis}
\label{sec:ablations}

Figure~\ref{fig:rm_analysis} shows that removing measured contact features degrades performance most clearly on \textit{Ground}. This indicates that contact information is important for motions with complex contact transitions. Adding future reference information performs slightly worse than the baseline, suggesting that this extra signal offers limited benefit and is difficult to exploit, it is sufficient to rely on the current and historical states to evaluate the quality of the action. Longer context improves alignment by revealing sliding, jitter, drift and recovery that isolated poses cannot capture.

Alignment also improves steadily as the available context grows from one to five seconds. Short segments capture instantaneous pose errors, whereas longer context reveals evolving artifacts such as foot sliding, repeated jitter, progressive drift, and recovery from
instability. HumanScore therefore benefits from integrating complementary evidence over time, rather than evaluating motion frame by frame.
\section{Conclusion}
\label{sec:conclusions}

We introduce a large-scale benchmark and HumanScore, a human-aligned metric, for evaluating humanoid motion tracking. Built from optical motion recorded from 24 professional performers, the released benchmark contains approximately 153 hours across four motion families. It reveals persistent weaknesses in highly dynamic and ground-contact motions.
\ours{} provides a reproducible framework for tracker comparison and failure analysis, with future work extending to cross-embodiment settings, real-world hardware, and reward optimization.

\bibliographystyle{assets/plainnat}
\bibliography{main}

\clearpage
\appendix
\balance
\section{Implementation details}

\subsection{Segment construction and padding}
Following Sec.~\ref{sec:pref}, a segment with $L<250$ frames is padded with zeros on the right to form a $250\times539$ input. A Boolean validity mask marks the first $L$ frames. The Transformer uses this mask to exclude padding from attention, and temporal mean pooling uses the same mask to exclude padding from the trajectory representation. For encoded token $h_t$ and feature index $j$, the pooled feature is
\[
\bar{h}_{j}=\frac{\sum_{t=1}^{250}m_t h_{t,j}}{\sum_{t=1}^{250}m_t},
\]
where $m_t\in\{0,1\}$. Short terminal segments can therefore contribute without padded frames affecting attention or pooling. When window rewards are combined across a rollout, each window is weighted by its number of actual frames. Padded positions contribute neither to the window representation nor to the trajectory average.

\subsection{Frame features}
Table~\ref{tab:rm_features} decomposes the reported 539-dimensional frame token into 70 current reference dimensions and 469 rollout dimensions. The rollout features are computed from the simulated robot. Foot contact and force are obtained from contacts between the robot and floor, while foot acceleration is the temporal derivative of measured foot velocity. Keypoint features describe 14 bodies in a navigation frame aligned with gravity.

\begin{table}[H]
\centering
\small
\caption{\textbf{HumanScore input features.} Dimensions are reported for one frame.}
\label{tab:rm_features}
\setlength{\tabcolsep}{3pt}
\begin{tabular}{p{0.25\linewidth}p{0.52\linewidth}r}
\toprule
\textbf{Group} & \textbf{Contents} & \textbf{Dim.} \\
\midrule
Current reference & root pose and navigation velocity; joint position and velocity; foot contact & 70 \\
Robot state and action & root and IMU pose; action; motor target; joint position and velocity & 126 \\
Measured contact dynamics & foot contact, force, velocity and acceleration & 20 \\
Root motion & pelvis and root velocities in local and navigation frames & 15 \\
Current keypoints & 14 $4\!\times\!4$ poses and six dimensional spatial velocities & 308 \\
\midrule
Total & concatenated frame token & 539 \\
\bottomrule
\end{tabular}
\end{table}

The model without measured contact features removes the 20-dimensional rollout contact block and has 519 input dimensions. The \emph{Kinematics Only} variant also removes the three-dimensional gravity frame angular velocity and has 516 dimensions. Both variants retain the current reference foot contact target.

\subsection{Reward model optimization}
The reported model uses a bidirectional Transformer with normalization before each sublayer and a reward head comprising three linear layers. Table~\ref{tab:supp_reward_hyperparams} lists the architecture and optimization settings recovered from the reported checkpoint. At inference, a sigmoid maps each unbounded window reward to a bounded reward between zero and one. HumanScore is 100 times their mean weighted by the number of actual frames in each window.

\begin{table}[H]
\centering
\small
\caption{\textbf{Settings of the reported HumanScore checkpoint.}}
\label{tab:supp_reward_hyperparams}
\setlength{\tabcolsep}{2pt}
\begin{tabular}{@{}>{\raggedright\arraybackslash}p{0.50\linewidth}>{\raggedright\arraybackslash}p{0.42\linewidth}@{}}
\toprule
\textbf{Hyperparameter} & \textbf{Value} \\
\midrule
Model dimension & 256 \\
Transformer layers & 4 \\
Attention heads & 8 \\
FFN dimension & 1024 \\
Pooling & masked mean \\
Maximum sequence length & 250 frames \\
Temporal downsampling & none \\
Batch size & 8 \\
Optimizer & AdamW \\
Learning rate & $1\times10^{-4}$ \\
Training epochs & 20 \\
Learning rate schedule & cosine with 10\% warmup \\
Maximum gradient norm & 1.0 \\
Dropout & 0.1 \\
Weight decay & $1\times10^{-5}$ \\
Preference temperature & 1.0 \\
Similar pair weight & 1.0 \\
Training precision & float32 \\
Random seed & 42 \\
Padding & right zero padding with a validity mask \\
\bottomrule
\end{tabular}
\end{table}

\section{Additional related work}

\subsection{Evaluation from human preferences}
Pairwise judgments are useful when people can recognize the quality of a structured output more readily than researchers can express it as a fixed analytic objective. Learning from human preferences~\cite{RLHF17} established this approach for reward learning, while InstructGPT~\cite{InstructGPT22} and iterative RLHF~\cite{IterativeRLHF24} scaled it for language model alignment. In human motion, InstructMotion~\cite{instructmotion24} uses comparisons to refine motion generation from text. MotionCritic~\cite{motioncritic25} learns a quality metric from MotionPercept, and the Motion Turing Test~\cite{motionturing26} measures whether humanoid motion appears human from kinematic observations. These methods evaluate generated motion or general human likeness. Our setting instead compares the physical quality of robot rollouts that follow the same reference.

Robotic reward learning provides a second line of related work. LIV~\cite{liv23} learns dense rewards from videos without actions, and RoboCLIP~\cite{roboclip23} derives rewards from video or text demonstrations. RoboReward~\cite{roboreward26} trains vision and language reward models on large robot datasets. Robo-Dopamine~\cite{robodopamine25} models manipulation progress from multiple views, while Robometer~\cite{robometer26} combines progress supervision with trajectory comparisons. These methods primarily support task completion or policy optimization. HumanScore instead evaluates humanoid motion tracking by integrating pose, contact, force and velocity evidence over synchronized rollouts of the same reference.

\section{Preference data statistics}

The preference catalogue uses only motions from the \ours{} training set. We divide the aligned tracker rollouts for every source motion into consecutive clips and retain the final shorter clip. After sorting the complete catalogue by family, source motion and temporal position, we select uniformly spaced clips. A balanced schedule assigns one of the six tracker combinations to each selected clip, and the corresponding two rollouts form a comparison.
Six doctoral researchers specializing in humanoid robotics annotated 6,000 original trajectory pairs. We bilaterally mirrored each pair, yielding 12,000 preference records for model development. Mirrored variants remain in the same source-motion partition, and preference alignment is evaluated only on the original unmirrored test comparisons.

The label set contains strict preferences, \emph{Similar} judgments and \emph{Cannot compare} judgments. We exclude \emph{Cannot compare} from reward model optimization. Records are split 80/20 by source \texttt{motion\_id} using seed 42, so clips from one motion cannot cross partitions. The split balances motion family, tracker pair, label, annotator, clip length and the number of sampled clips from each source motion. Alignment analysis uses strict test samples. Agreement is computed within each family before the four family rates are averaged with equal weight. Samples too short to define acceleration or jerk are omitted for those metrics.

\section{Dataset scale and release format}
\label{sec:supp_release_integrity}

All released robot trajectories are stored at 50\,Hz. The training manifest lists 22,495 trajectories and 24,793,129 frames, while the test manifest lists 2,500 trajectories and 2,687,461 frames. This gives a 9:1 split by trajectory count. Table~\ref{tab:dataset_category_stats} reports the combined scale of each motion family.

\begin{table}[H]
\centering
\small
\caption{\textbf{\ours{} statistics by motion family.}}
\label{tab:dataset_category_stats}
\setlength{\tabcolsep}{3pt}
\begin{tabular}{lccc}
\toprule
\textbf{Family} & \textbf{Trajectories} & \textbf{Frames} & \textbf{Duration (h)} \\
\midrule
Daily & 9,739 & 16,072,017 & 89.29 \\
Highly Dynamic & 2,676 & 1,981,843 & 11.01 \\
Ground & 1,640 & 825,792 & 4.59 \\
Interaction & 10,940 & 8,600,938 & 47.78 \\
\midrule
Total & 24,995 & 27,480,590 & 152.67 \\
\bottomrule
\end{tabular}
\end{table}

The \texttt{train.json} and \texttt{test.json} manifests define the released partitions. All trajectories derived from the same source motion remain in one partition, preventing related entries from crossing between training and test data. Each manifest entry records its path, motion family and frame count.

Each NPZ archive contains seven arrays with a common temporal length $F$. The robot state arrays are \texttt{qpos} of shape $F\times36$ and \texttt{qvel} of shape $F\times35$, both in \texttt{float32}. Keypoint motion is stored as \texttt{kpt2gv\_pose} of shape $F\times14\times4\times4$ in \texttt{float32} and \texttt{kpt\_cvel\_in\_gv} of shape $F\times14\times6$ in \texttt{float64}. Global motion uses \texttt{gv\_vel} of shape $F\times3$ and \texttt{gv2wrd\_pose} of shape $F\times4\times4$, both in \texttt{float32}. The Boolean array \texttt{foot\_contact} has shape $F\times2$. A complete scan confirmed this schema across all 24,995 released archives.

\section{Analysis of motion coverage}
\label{sec:supp_motion_coverage}

We characterize each motion family using four descriptors computed from one source reference trajectory per motion. Horizontal root path is the accumulated root translation in the ground plane. Root height range is the difference between the maximum and minimum root height. For joint speed, we take finite differences of the 29 actuated joint coordinates at 50\,Hz and then compute the 95th percentile of absolute speed across frames and joints. Table~\ref{tab:supp_motion_descriptors} reports the median, first quartile and third quartile of each descriptor across source motions.

Daily motions are longer and cover more horizontal distance, whereas Interaction motions are spatially localized. Ground spans a broader upper half of the root height distribution. Highly Dynamic contains many short skills, so its label does not imply that every kinematic descriptor exceeds those of other families. These descriptors do not measure contact timing, support or recovery and therefore cannot replace Succ, MPJPE or HumanScore.

\begin{table*}[t]
\centering
\small
\caption{\textbf{Kinematic coverage of \ours{} reference motions.} Each entry reports median [first quartile, third quartile] across source motions.}
\label{tab:supp_motion_descriptors}
\setlength{\tabcolsep}{5pt}
\resizebox{0.98\linewidth}{!}{
\begin{tabular}{lcccc}
\toprule
\textbf{Family} & \textbf{Duration (s)} & \textbf{Horizontal root path (m)} & \textbf{Root height range (m)} & \textbf{Joint speed P95 (rad/s)} \\
\midrule
Daily & 30.81 [27.62, 33.68] & 15.25 [10.03, 20.99] & 0.273 [0.108, 0.428] & 2.316 [1.865, 2.770] \\
Highly Dynamic & 8.38 [6.70, 11.17] & 1.40 [0.63, 2.69] & 0.081 [0.045, 0.114] & 1.657 [1.221, 2.051] \\
Interaction & 10.24 [8.58, 11.70] & 0.42 [0.29, 1.16] & 0.008 [0.005, 0.045] & 0.962 [0.626, 1.157] \\
Ground & 9.97 [9.14, 11.14] & 0.66 [0.14, 1.20] & 0.102 [0.087, 0.620] & 1.079 [0.843, 1.255] \\
\bottomrule
\end{tabular}
}
\end{table*}

\section{Analysis of preference alignment}
\label{sec:supp_alignment_uncertainty}

Table~\ref{tab:supp_alignment_ci} supplements Table~\ref{tab:preference_alignment} with uncertainty for Align Rate. We use 20,000 bootstrap draws with seed 20260813. Within each family, source motions are sampled with replacement and all associated comparisons are retained. We then recompute the four family rates and their equal-weight mean. This procedure preserves the reported family weighting while accounting for repeated comparisons from the same source motion.

\begin{table}[H]
\centering
\small
\caption{\textbf{Uncertainty of preference alignment.} We report 95\% intervals from a bootstrap stratified by motion family and clustered by source motion.}
\label{tab:supp_alignment_ci}
\setlength{\tabcolsep}{4pt}
\resizebox{\linewidth}{!}{
\begin{tabular}{lrr}
\toprule
\textbf{Metric} & \textbf{Align Rate (\%)} & \textbf{95\% CI (\%)} \\
\midrule
HumanScore & 90.83 & [87.36, 93.83] \\
MPJPE & 80.49 & [75.95, 84.76] \\
MPJVE & 84.04 & [79.80, 87.87] \\
KPT Position MAE & 84.05 & [79.67, 88.04] \\
Foot Contact Accuracy & 78.82 & [73.73, 83.59] \\
Avg Joint Accel & 69.33 & [64.23, 74.12] \\
Avg Joint Jerk & 72.32 & [67.52, 76.93] \\
\bottomrule
\end{tabular}
}
\end{table}

\section{Discussion and Limitations}
\label{sec:limitations}

The benchmark and preference results point to the same conclusion from different directions. Category-level evaluation shows that contact regime changes the relative behaviour of trackers: a method that is reliable on upright daily motion can still fail almost completely during ground-level transitions. The preference experiment explains why this distinction is not fully represented by joint error. HumanScore gains most of its advantage when it can observe contact-related state over several seconds, suggesting that the perceptual unit of failure is often an event such as a slide, impact, support switch or recovery, rather than an isolated pose. HumanScore should therefore be read alongside success and kinematic error, not as a replacement for all analytic diagnostics: the former summarizes perceived trajectory quality, whereas the latter remain valuable for locating a specific source of error.

Several boundaries remain. HumanScore is trained from \ours{} training motions and rollouts produced by four trackers, so its motion-disjoint test measures generalization to unseen motions, not to an entirely unseen robot, simulator or controller family. Its 539-dimensional input also includes privileged simulator state and contact quantities that are available for benchmarking but may not be observable on hardware; applying the metric to real-world trajectories will require an observable feature set or a separately validated state estimator. The preference pool assigns one primary judgment to each pair, which provides broad coverage but does not quantify uncertainty through repeated independent labels. Finally, the present metric comparison covers representative individual diagnostics; broader tests against fitted linear and nonlinear diagnostic composites would further delimit what must be learned from trajectory-level preference data.

\ours{} also inherits the limits of its data distribution. The four families are deliberately diagnostic but imbalanced, with far fewer Ground clips than Daily or Interaction clips, and the evaluation uses one 29-DoF humanoid embodiment in MuJoCo. Results should not be interpreted as a population estimate over all human activity or as evidence of hardware robustness. Extending the benchmark to additional embodiments, real-robot rollouts and rare contact regimes is therefore a natural next step. We also restrict HumanScore to evaluation in this work. Directly optimizing a learned score can create behaviours that exploit model imperfections; using it as a reinforcement-learning reward will require explicit regularization and an independent human evaluation rather than assuming that benchmark alignment transfers automatically to policy optimization.

\end{document}